\documentclass[conference]{IEEEtran}
\IEEEoverridecommandlockouts
\usepackage{cite}
\usepackage{amsmath,amssymb,amsfonts}
\usepackage{algorithmic}
\usepackage{graphicx}
\usepackage{textcomp}
\usepackage{xcolor}
\usepackage{booktabs}
\usepackage{multirow}
\usepackage{booktabs} 
\usepackage{graphicx} 
\usepackage[colorlinks=true, linkcolor=blue, citecolor=blue, urlcolor=blue]{hyperref}
\usepackage{xcolor}
\def\BibTeX{{\rm B\kern-.05em{\sc i\kern-.025em b}\kern-.08em
    T\kern-.1667em\lower.7ex\hbox{E}\kern-.125emX}}
\begin{document}

\title{ReLaMix: Residual Latency-Aware Mixing for Delay-Robust Financial Time-Series Forecasting\\
}

\author{
\IEEEauthorblockN{
Tianyou Lai\textsuperscript{1}\textsuperscript{\textdagger},
Wentao Yue\textsuperscript{1}\textsuperscript{\textdagger},
Jiayi Zhou\textsuperscript{1}\textsuperscript{\textdagger},
Chaoyuan Hao\textsuperscript{1}, \\
Lingke Chang\textsuperscript{2},
Qingyu Mao\textsuperscript{3},
Zhibo Niu\textsuperscript{4},
Qilei Li\textsuperscript{5}\textsuperscript{*}
}
\IEEEauthorblockA{
\textsuperscript{1}Lanzhou University, Lanzhou, China \\
\textsuperscript{2}Brandeis University, Waltham, USA \\
\textsuperscript{3}Shenzhen University, Shenzhen, China \\
\textsuperscript{4}Central University of Finance and Economics, Beijing, China \\
\textsuperscript{5}Central China Normal University, Wuhan, China
}
}

\maketitle
\begingroup
\renewcommand\thefootnote{}
\footnotetext{\textsuperscript{\textdagger}Co-first authors contributed equally.}
\footnotetext{\textsuperscript{*}Corresponding author: qilei.li@ccnu.edu.cn}
\endgroup

\begin{abstract}
Financial time-series forecasting in real-world high-frequency markets is often hindered by delayed or partially stale observations caused by asynchronous data acquisition and transmission latency. To better reflect such practical conditions, we investigate a simulated delay setting where a portion of historical signals is corrupted by a Zero-Order Hold (ZOH) mechanism, significantly increasing forecasting difficulty through stepwise stagnation artifacts. In this paper, we propose ReLaMix (Residual Latency-Aware Mixing Network), a lightweight extension of TimeMixer that integrates learnable bottleneck compression with residual refinement for robust signal recovery under delayed observations. ReLaMix explicitly suppresses redundancy from repeated stale values while preserving informative market dynamics via residual mixing enhancement. Experiments on a large-scale second-resolution PAXGUSDT benchmark demonstrate that ReLaMix consistently achieves state-of-the-art accuracy across multiple delay ratios and prediction horizons, outperforming strong mixer and Transformer baselines with substantially fewer parameters. Moreover, additional evaluations on BTCUSDT confirm the cross-asset generalization ability of the proposed framework. These results highlight the effectiveness of residual bottleneck mixing for high-frequency financial forecasting under realistic latency-induced staleness.
\end{abstract}


\section{Introduction}

\textit{``The future is already here — it’s just not very evenly distributed."}— William Gibson

High-frequency trading (HFT) is central to modern financial markets, where predictive algorithms increasingly depend on real-time streams of raw Limit Order Book (LOB) signals to capture microstructure dynamics~\cite{moreno2024deepvol, zhang2019deeplob}. 
Despite the common assumption of continuously available observations, practical trading systems operate under unavoidable physical constraints~\cite{briola2025deep}. 
Network congestion, hardware latency, and exchange interface bottlenecks frequently disrupt data transmission, producing stale and asynchronously updated price signals~\cite{kong2025staleness}. 
Such disruptions often manifest as Zero-Order Hold (ZOH) stagnation, where the observed values remain constant until new updates arrive. 
Importantly, this phenomenon introduces statistical bias rather than benign technical noise: recent evidence shows that staleness can be misidentified as artificial price jumps or spuriously low volatility, undermining the reliability of downstream forecasting models~\cite{kolokolov2024jumps, kong2025staleness, park2025reliable}.

\begin{figure}[t]
    \centering
    \includegraphics[width=\linewidth]{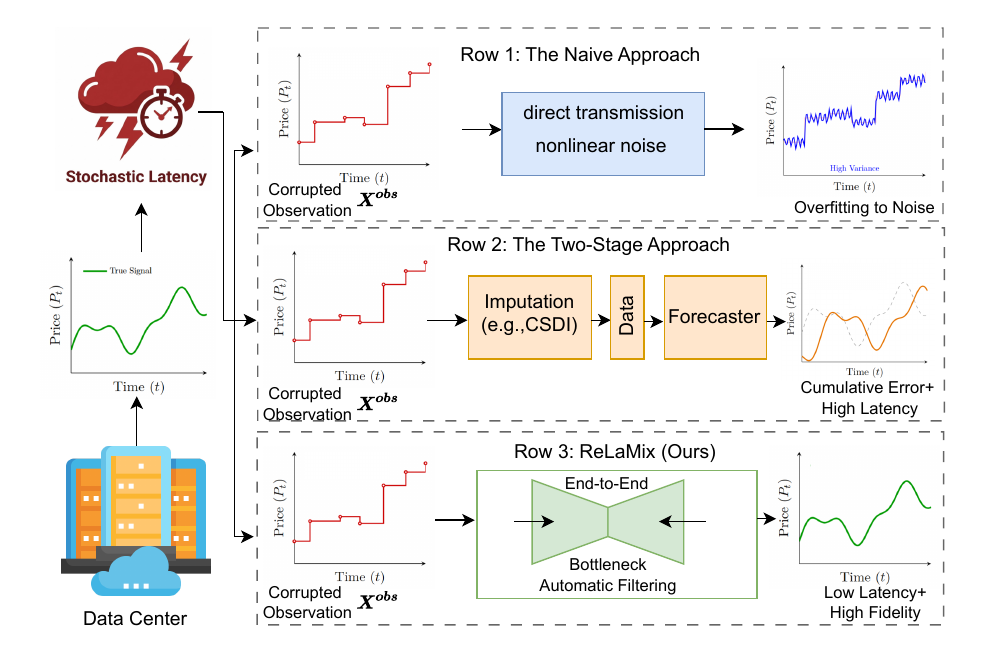}
    \caption{Comparison of High-Frequency Signal Recovery Paradigms. 
    Row 1: Naive models overfit ZOH noise. 
    Row 2: Two-stage pipelines introduce additional latency. 
    Row 3 (Ours): ReLaMix employs an end-to-end bottleneck mixer to efficiently recover clean signals.}
    \label{fig:workflow}
\end{figure}

Unlike explicit missing values (e.g., NaNs) considered in standard imputation settings~\cite{tashiro2021csdi, yang2025revisiting}, ZOH corruption presents as repetitions of numerically valid observations, forming step-function artifacts that conceal true market fluctuations~\cite{moreno2024deepvol}. 
As illustrated in Fig.~\ref{fig:workflow}, recovering clean high-frequency dynamics under such latency-induced stagnation requires models that can suppress redundancy while remaining computationally efficient for real-time deployment.

Existing deep forecasting architectures face distinct limitations in this regime. 
Sequential recurrent models such as LSTM~\cite{hochreiter1997long} and GRU~\cite{dey2017gate} lack global receptive fields to disentangle ZOH artifacts from meaningful temporal dynamics. 
Meanwhile, long-horizon predictors—including convolution-based architectures such as TimesNet~\cite{wu2022timesnet} and attention-based Transformers like PatchTST~\cite{nie2022time} and LiT~\cite{xiao2025lit}—excel in long-range modeling but incur substantial computational overhead for short high-frequency recovery tasks~\cite{xu2023fits}.
This complexity leads to unacceptable inference latency in HFT environments. 
Even efficient mixer-style designs such as TimeMixer~\cite{wang2024timemixer} remain limited, as pooling-based downsampling can oversmooth sharp ZOH step-functions, causing critical information loss. 
Moreover, over-parameterized networks tend to memorize stagnation patterns and underestimate latent volatility, resulting in poor generalization under delayed observations~\cite{yang2025revisiting,liu2022non}.

To bridge the gap between recovery accuracy and deployment efficiency, we propose \emph{ReLaMix}, a lightweight Residual Latency-Aware Mixing framework. 
Grounded in the Information Bottleneck principle~\cite{alemi2016deep}, ReLaMix introduces an explicit Bottleneck Compression module to filter ZOH-induced redundancy by projecting noisy inputs into a compact latent space~\cite{li2023ti}. 
To preserve expressive power, we further integrate an Expand--Compress strategy within a pre-activation residual mixing block, enabling the model to capture nonlinear market dynamics with minimal parameter overhead~\cite{yue2025modularity}. 
This design supports a ``less-is-more'' paradigm for high-frequency forecasting under latency constraints~\cite{chen2023tsmixer, zeng2023transformers}.

The main contributions are summarized as follows:
\begin{itemize}
    \item We formalize high-frequency forecasting under ZOH stagnation, and reveal the key failure modes of existing deep forecasting architectures, including noise overfitting and computational redundancy under delayed observations.
    
    \item We propose \textbf{ReLaMix}, a principled and lightweight residual bottleneck mixing network that explicitly suppresses latency-induced staleness while preserving nonlinear market dynamics for robust signal recovery.
    
    \item We design a strictly non-overlapping segmented evaluation protocol to prevent temporal leakage, and demonstrate state-of-the-art performance on second-resolution PAXGUSDT forecasting benchmarks, with consistent generalization to BTCUSDT and substantially fewer parameters than strong mixer and Transformer baselines.
\end{itemize}

\section{Related Work}

High-frequency financial forecasting has shifted from aggregated daily indicators to raw intraday streams to better capture market microstructure~\cite{moreno2024deepvol}. 
Such high-resolution observations are inevitably affected by asynchrony and latency, often producing ZOH-type staleness. 
Recent econometrics studies emphasize that staleness constitutes a distinct source of bias rather than simple missingness: Kolokolov et al.~\cite{kolokolov2024jumps} showed that untreated stagnation may be misinterpreted as price jumps, while Kong et al.~\cite{kong2025staleness} introduced factor-based quantification of this friction. 
Compared to these assumption-driven statistical treatments, we pursue an efficient deep learning approach for signal recovery under delayed observations.

Handling irregular or delayed measurements is a long-standing challenge in time-series learning. 
Classical pipelines typically rely on imputation, including diffusion-based models such as CSDI~\cite{tashiro2021csdi}. 
However, Yang et al.~\cite{yang2025revisiting} argued that imputation-then-prediction can induce distribution shifts, motivating end-to-end learning through the Information Bottleneck principle~\cite{alemi2016deep}. 
Continuous-time formulations, such as Neural Controlled Differential Equations~\cite{kidger2020neural}, provide theoretical flexibility but remain computationally costly for latency-sensitive forecasting. 
ReLaMix follows the end-to-end philosophy of~\cite{yang2025revisiting} while adopting a substantially lighter architecture tailored for real-time inference.

Meanwhile, recent work has discussed potential limitations of Transformer-based forecasting models in terms of computational overhead and inductive suitability~\cite{zeng2023transformers}, motivating efficient MLP-style mixers such as TSMixer~\cite{chen2023tsmixer} and TimeMixer~\cite{wang2024timemixer}. 
However, these architectures are not explicitly designed for latency-corrupted inputs: pooling-based decomposition may blur sharp ZOH-induced stagnation and genuine low-volatility dynamics. 
In contrast, ReLaMix employs learnable bottleneck compression to suppress redundancy from repeated observations, combined with residual refinement to preserve informative market signals, while retaining mixer-level efficiency.

\section{METHODOLOGY}
\begin{figure}[t]
    \centering
    \includegraphics[width=\columnwidth,height=0.99\textheight,keepaspectratio]{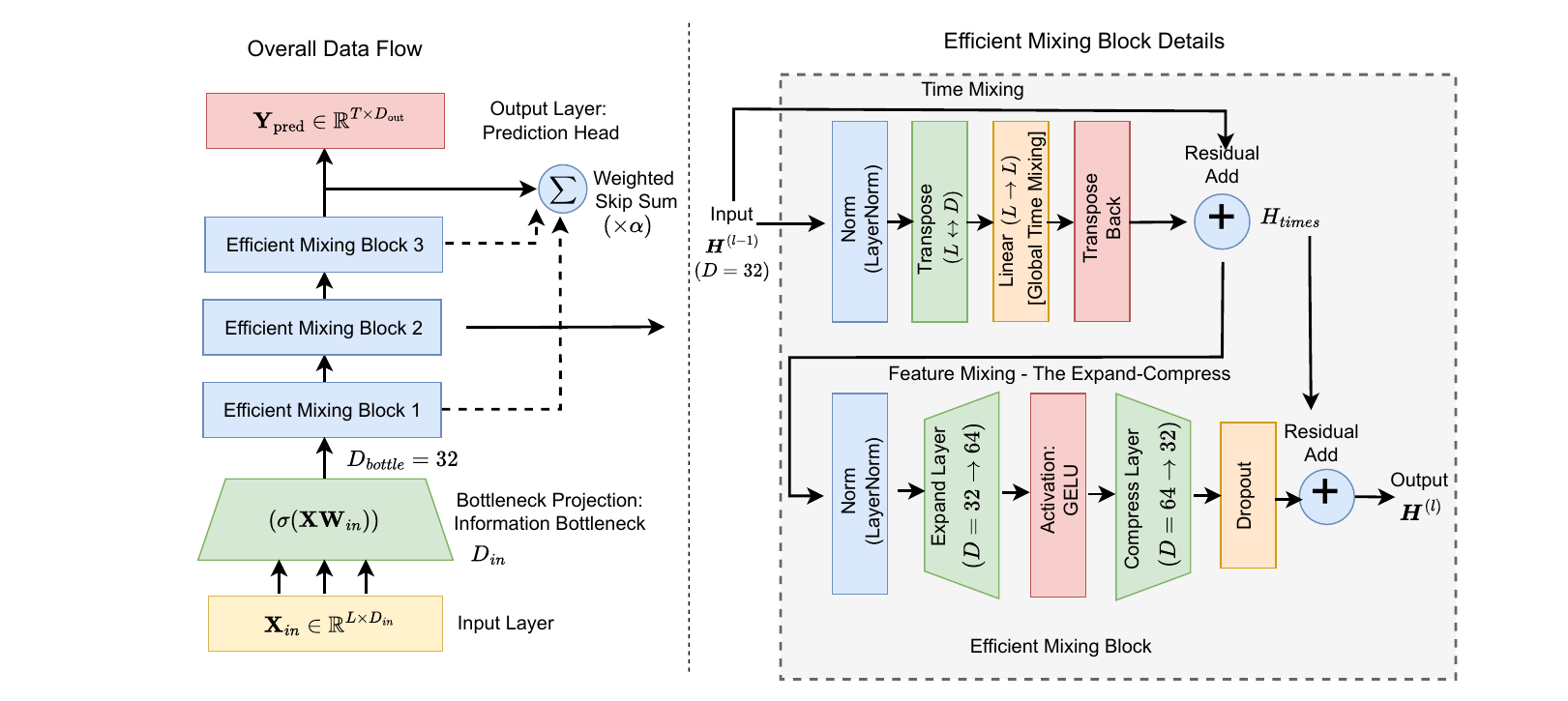}
    \caption{Overview of ReLaMix. The model combines an information bottleneck projection with stacked mixing blocks and dense residual refinement for robust forecasting under ZOH-delayed observations.}
    \label{fig:ReLaMix_arch}
\end{figure}

In this section, we formalize the signal recovery problem under ZOH latency constraints and introduce \textbf{ReLaMix} (Residual Latency-Aware Mixing Network), a lightweight framework for forecasting with delayed observations. 
As illustrated in Fig.~\ref{fig:ReLaMix_arch}, ReLaMix first compresses high-dimensional corrupted inputs into a compact bottleneck space ($D_{in}\!\to\!32$), and then progressively refines representations through stacked Efficient Mixing Blocks and multi-level skip connections. 
Each block couples temporal mixing with an Expand--Compress feature pathway ($32\!\to\!64\!\to\!32$) to suppress ZOH-induced artifacts while preserving nonlinear market dynamics. 
We detail these components below.

\subsection{Problem Formulation and ZOH Modeling}

Let $\mathbf{X}=\{\mathbf{x}_1,\mathbf{x}_2,\dots,\mathbf{x}_T\}$ denote the underlying clean second-resolution financial time series (ground truth), where $\mathbf{x}_t\in\mathbb{R}^{D_{in}}$ is the multivariate market feature vector at time step $t$, including standard OHLCV (Open, High, Low, Close, and Volume) indicators. 

In realistic high-frequency trading environments, observations are often corrupted by stochastic transmission latency, producing delayed or stale inputs. We denote the observed sequence as $\mathbf{X}^{obs}$, generated from $\mathbf{X}$ via a Zero-Order Hold (ZOH) mechanism. Specifically, we introduce a binary latent state $s_t\in\{0,1\}$, where $s_t=0$ indicates a stagnation event (no update received) and $s_t=1$ denotes a valid observation update.

Accordingly, the observed value at time step $t$ is recursively defined as
\vspace{-0.6em}
\begin{equation}
    \mathbf{x}^{obs}_t = 
    \begin{cases} 
    \mathbf{x}_t & \text{if } s_t = 1 \\
    \mathbf{x}^{obs}_{t-1} & \text{if } s_t = 0
    \end{cases}
\end{equation}
\vspace{-0.1em}
As shown in Eq.~(1), ZOH latency produces step-function artifacts, where observed prices remain artificially constant during stagnation intervals. Such stale signals obscure true market fluctuations and substantially increase forecasting difficulty.

Given a corrupted historical window $\mathbf{X}^{obs}_{t-L+1:t}$ of length $L$, our goal is to learn a parameterized predictor $f_{\theta}(\cdot)$ that robustly recovers the future clean market state at horizon $k$, i.e.,

\vspace{-1.8em}
\begin{equation}
\hat{\mathbf{x}}_{t+k} = f_\theta!\left(\mathbf{X}^{obs}_{t-L+1:t}\right),
\vspace{-0.5em}
\end{equation}

despite the presence of ZOH-induced information staleness. This formulation defines a delay-robust financial signal recovery and forecasting problem, which serves as the foundation of our proposed ReLaMix framework.

\subsection{Information Bottleneck Compression}

Standard deep forecasting models often encode inputs into high-dimensional latent spaces to capture complex temporal dependencies. However, under ZOH corruption, the observed stream $\mathbf{X}^{obs}$ contains substantial redundancy: latency-induced stagnation produces repeated stepwise artifacts that can dominate representation learning. Directly projecting such inputs into an over-parameterized space may thus encourage overfitting to stale patterns rather than recovering informative market dynamics.

To mitigate this issue, ReLaMix adopts the Information Bottleneck principle by compressing features into a low-dimensional bottleneck space. This explicit compression filters task-irrelevant ZOH noise while retaining the most predictive components for forecasting.

Formally, given an observed window $\mathbf{X}^{obs} \in \mathbb{R}^{L \times D_{in}}$, we apply a learnable bottleneck projection:

\vspace{-0.8em}
\begin{equation}
\vspace{-0.6em}
\mathbf{H}^{(0)} = \sigma\!\left(\mathbf{X}^{obs}\mathbf{W}_{in} + \mathbf{b}_{in}\right),
\vspace{-0.1em}
\end{equation}

where $\mathbf{W}_{in} \in \mathbb{R}^{D_{in} \times d_b}$ is the compression matrix, $\mathbf{b}_{in}$ is the bias, $\sigma(\cdot)$ denotes a nonlinear activation (e.g., GELU), and $d_b \ll D_{in}$ is the bottleneck dimension. In practice, we set $d_b=\tfrac{1}{2}D_{model}$ for a lightweight design.

This compact encoding suppresses ZOH-induced repeated-value artifacts, reducing overfitting to stale observations and preserving predictive signals before subsequent temporal mixing and residual refinement.

\subsection{Residual Latency-Aware Mixing Block}

The core component of ReLaMix is the proposed Residual Latency-Aware Mixing Block, which serves as the fundamental building module for robust representation learning under ZOH-corrupted observations. To balance computational efficiency with sufficient representational capacity, we adopt a lightweight pre-activation (pre-norm) residual design, where normalization is applied before each mixing operation. Each block consists of two complementary sub-modules: Time Mixing, which captures temporal dependencies along the sequence dimension, and Feature Mixing, which models cross-variable interactions through an efficient bottleneck expansion–compression mechanism.

\subsubsection{Time Mixing: Capturing Temporal Dependencies}

The Time Mixing module is designed to capture temporal dependencies from the compressed bottleneck representations. Given the input hidden state $\mathbf{H}^{(l-1)} \in \mathbb{R}^{L \times d_b}$ at layer $l-1$, we adopt a pre-normalization scheme to stabilize training:

\vspace{-0.8em}
\begin{equation}
\tilde{\mathbf{H}} = \mathrm{LayerNorm}\!\left(\mathbf{H}^{(l-1)}\right).
\vspace{-0.6em}
\end{equation}

To explicitly model interactions across time steps, ReLaMix performs linear mixing along the temporal dimension. Specifically, we transpose the sequence representation and apply a learnable temporal transformation:
\vspace{-0.8em}
\begin{equation}
\mathbf{H}_{temp} =
\left(\tilde{\mathbf{H}}^{\top}\mathbf{W}_{t} + \mathbf{b}_{t}\right)^{\top},
\vspace{-0.6em}
\end{equation}

where $\mathbf{W}_{t} \in \mathbb{R}^{L \times L}$ denotes the temporal mixing matrix shared across feature channels, and $\mathbf{b}_{t}$ is the bias term. In practice, this operation is implemented using a lightweight linear layer followed by nonlinearity and dropout for regularization.

Finally, we add a residual connection to preserve the signal and improve robustness to latency-induced staleness:

\vspace{-0.8em}
\begin{equation}
\mathbf{H}^{(l)} = \mathbf{H}^{(l-1)} + \mathbf{H}_{temp}.
\vspace{-0.6em}
\end{equation}

This temporal mixing operation enables the model to aggregate historical information across multiple time steps, effectively smoothing local discontinuities introduced by the ZOH mechanism and recovering informative market dynamics.

\subsubsection{Feature Mixing: Expand--Compress Mechanism}

After Time Mixing aggregates temporal dependencies, the model must further capture nonlinear interactions among financial variables (e.g., price--volume coupling). Conventional residual blocks often preserve a constant high-dimensional width, which can introduce parameter redundancy under ZOH-corrupted inputs. To balance efficiency and expressiveness, ReLaMix adopts an \emph{expand--compress} feature mixing mechanism, analogous to inverted residual designs in lightweight architectures.

Formally, given the temporally mixed representation $\mathbf{H}_{time}$, we first apply pre-normalization:

\vspace{-0.8em}
\begin{equation}
\tilde{\mathbf{H}} = \mathrm{LayerNorm}\!\left(\mathbf{H}_{time}\right).
\vspace{-0.6em}
\end{equation}

We then expand the bottleneck features into a higher-dimensional latent space to enhance interaction modeling:

\vspace{-0.8em}
\begin{equation}
\mathbf{H}_{exp} = \sigma\!\left(\tilde{\mathbf{H}}\mathbf{W}_{exp} + \mathbf{b}_{exp}\right),
\vspace{-0.6em}
\end{equation}
\begin{figure}[t]
    \centering
    \includegraphics[width=\columnwidth]{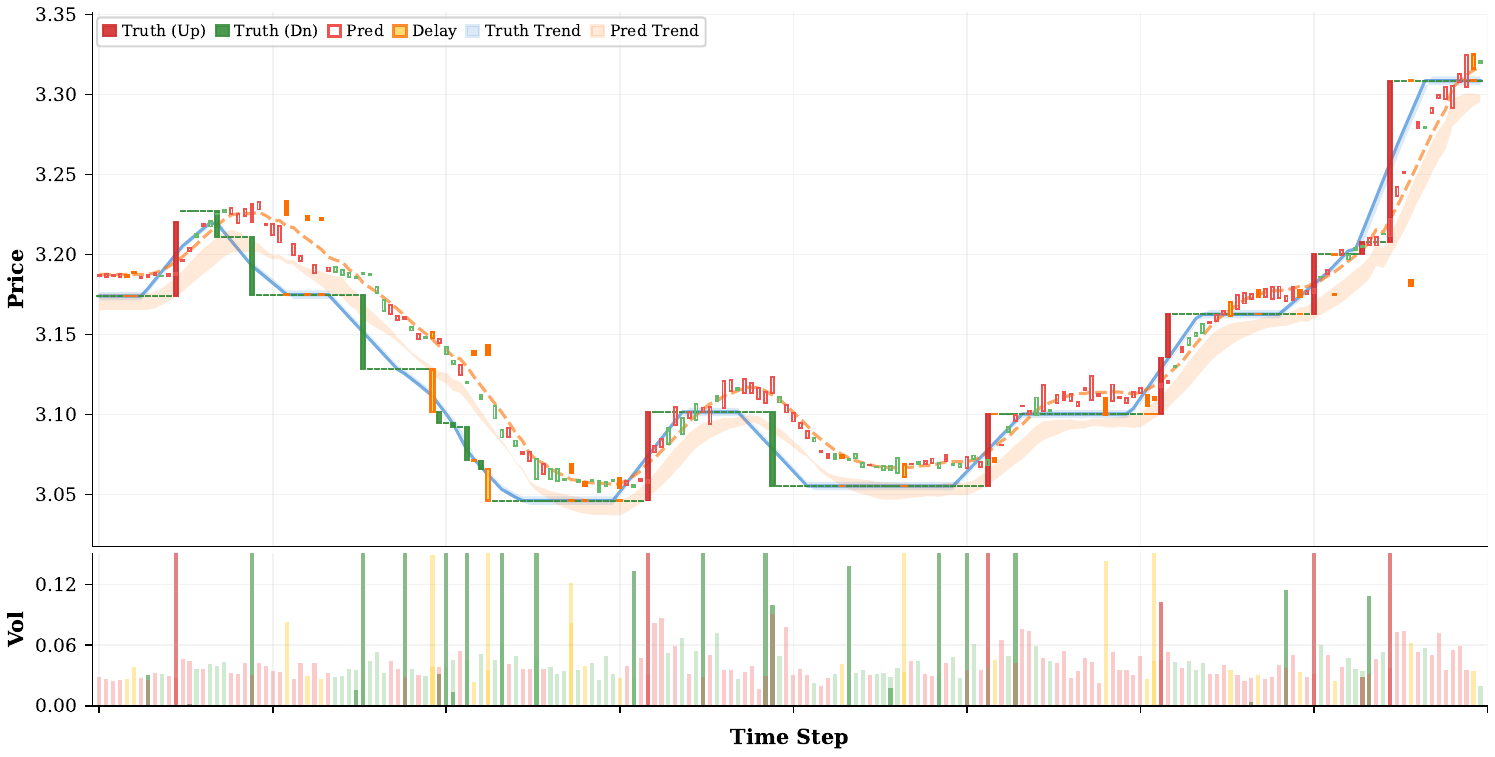}
    
    \caption{Qualitative visualization of ReLaMix signal recovery on PAXGUSDT. ReLaMix restores high-frequency K-line dynamics and trading volume variations under simulated ZOH delays.}
    
    \label{fig:kline_volume}
\end{figure}
followed by a compression step that projects the enriched representation back to the bottleneck dimension:

\vspace{-1.2em}
\begin{equation}
\mathbf{H}_{comp} = \mathbf{H}_{exp}\mathbf{W}_{comp} + \mathbf{b}_{comp}.
\vspace{-0.8em}
\end{equation}

Finally, a residual connection is introduced to preserve the original signal:

\vspace{-1.2em}
\begin{equation}
\mathbf{H}^{(l)} = \mathbf{H}_{time} + \mathbf{H}_{comp}.
\vspace{-1em}
\end{equation}

Here, $\mathbf{W}_{exp} \in \mathbb{R}^{d_b \times D_{model}}$ and $\mathbf{W}_{comp} \in \mathbb{R}^{D_{model} \times d_b}$ denote the expansion and compression matrices, respectively. In our implementation, we set $d_b=32$ and $D_{model}=64$, corresponding to an intermediate expansion from $32 \rightarrow 64 \rightarrow 32$. Nonlinear activation (e.g., GELU) and dropout are applied for regularization in practice.

This expand--compress design ensures that ReLaMix maintains sufficient capacity to disentangle informative market dynamics from ZOH-induced staleness, while only propagating compact representations across layers, consistent with the Information Bottleneck principle.

\begin{table*}[!tbp]
\centering
\caption{SOTA comparison and ablation results on \textbf{PAXGUSDT} under simulated ZOH delays. 
The best and second-best results are highlighted in \textcolor{red}{red} 
and \textcolor{blue}{blue}, respectively. Ablation variants are excluded from 
the ranking.}
\label{tab:sota_ablation_paxg_metric}
\setlength{\tabcolsep}{3pt}
\renewcommand{\arraystretch}{1.08}
\resizebox{\textwidth}{!}{
\begin{tabular}{l|c|c|cccc|cccc|cccc}
\toprule
\multirow{2}{*}{\textbf{Model}} &
\multirow{2}{*}{\textbf{Type}} &
\multirow{2}{*}{\textbf{Metric}} &
\multicolumn{4}{c|}{\textbf{Delay 15\%}} &
\multicolumn{4}{c|}{\textbf{Delay 25\%}} &
\multicolumn{4}{c}{\textbf{Delay 35\%}} \\
\cline{4-15}
& & &
$k{=}1$ & $k{=}5$ & $k{=}7$ & $k{=}10$ &
$k{=}1$ & $k{=}5$ & $k{=}7$ & $k{=}10$ &
$k{=}1$ & $k{=}5$ & $k{=}7$ & $k{=}10$ \\
\midrule

\multirow{4}{*}{LSTM~\cite{hochreiter1997long}} & \multirow{4}{*}{RNN}
& MSE &
0.73306 & 1.003 & 1.073 & 1.194 &
0.74987 & 1.076 & 1.12 & 1.112 &
0.70363 & 1.166 & 1.176 & 1.094 \\
& & MAE &
0.61313 & 0.69925 & 0.78572 & 0.82405 &
0.65159 & 0.72386 & 0.81053 & 0.78591 &
0.61939 & 0.76006 & 0.84618 & 0.7905 \\
& & $R^2$ &
0.77519 & 0.70017 & 0.67701 & 0.6388 &
0.77004 & 0.67859 & 0.66273 & 0.66387 &
0.78422 & 0.65142 & 0.64606 & 0.66352 \\
& & Params &
51,781 & 53,081 & 53,731 & 54,706 &
51,781 & 53,081 & 53,731 & 54,706 &
51,781 & 53,081 & 53,731 & 54,706 \\
\midrule

\multirow{4}{*}{GRU~\cite{dey2017gate}} & \multirow{4}{*}{RNN}
& MSE &
0.44374 & 0.5643 & 0.57861 & 0.65081 &
0.40418 & 0.6381 & 0.60754 & 0.63042 &
0.51218 & 0.59574 & 0.68202 & 0.72566 \\
& & MAE &
0.45089 & 0.53354 & 0.60384 & 0.61127 &
0.44136 & 0.57014 & 0.63112 & 0.61941 &
0.49958 & 0.54633 & 0.64548 & 0.63309 \\
& & $R^2$ &
0.86392 & 0.82695 & 0.82257 & 0.80043 &
0.87605 & 0.80433 & 0.8137 & 0.80669 &
0.84294 & 0.81734 & 0.79088 & 0.77751 \\
& & Params &
\textcolor{blue}{38,917} & \textcolor{blue}{40,217} & \textcolor{blue}{40,867} & \textcolor{blue}{41,842} &
\textcolor{blue}{38,917} & \textcolor{blue}{40,217} & \textcolor{blue}{40,867} & \textcolor{blue}{41,842} &
\textcolor{blue}{38,917} & \textcolor{blue}{40,217} & \textcolor{blue}{40,867} & \textcolor{blue}{41,842} \\
\midrule

\multirow{4}{*}{TimeMixer~\cite{wang2024timemixer}} & \multirow{4}{*}{Mixer}
& MSE &
0.07898 & 0.09565 & 0.12284 & 0.15056 &
0.09318 & 0.13007 & 0.13632 & \textcolor{blue}{0.14141} &
0.13169 & 0.13601 & 0.15294 & 0.19659 \\
& & MAE &
0.20345 & 0.2084 & 0.23464 & 0.26635 &
\textcolor{blue}{0.21755} & \textcolor{blue}{0.2397} & \textcolor{blue}{0.24684} & \textcolor{blue}{0.24445} &
0.26531 & 0.24685 & \textcolor{blue}{0.26825} & 0.29751 \\
& & $R^2$ &
0.97578 & 0.97067 & 0.96234 & 0.95384 &
0.97143 & 0.96012 & 0.95821 & \textcolor{blue}{0.95665} &
0.95963 & 0.95831 & 0.95312 & 0.93975 \\
& & Params &
41,293 & 41,953 & 42,283 & 42,778 &
41,293 & 41,953 & 42,283 & 42,778 &
41,293 & 41,953 & 42,283 & 42,778 \\
\midrule

\multirow{4}{*}{TimesNet~\cite{wu2022timesnet}} & \multirow{4}{*}{CNN}
& MSE &
\textcolor{blue}{0.04904} & 0.07108 & 0.10347 & \textcolor{blue}{0.10499} &
\textcolor{blue}{0.07111} & 0.10615 & 0.11821 & 0.14739 &
0.08286 & \textcolor{blue}{0.10134} & 0.14467 & 0.18356\\
& & MAE &
\textcolor{blue}{0.19467} & 0.21985 & 0.25941 & \textcolor{blue}{0.2419} &
0.22219 & 0.25151 & 0.26074 & 0.27065 &
0.23656 & \textcolor{blue}{0.24434} & 0.28872 & \textcolor{blue}{0.28923} \\
& & $R^2$ &
\textcolor{blue}{0.98496} & 0.9782 & 0.96826 & \textcolor{blue}{0.9678} &
\textcolor{blue}{0.97819} & 0.96745 & 0.96376 & 0.95483 &
0.97459 & \textcolor{blue}{0.96894} & 0.95566 & 0.94658 \\
& & Params &
228,741 & 229,401 & 229,731 & 230,226 &
228,741 & 229,401 & 229,731 & 230,226 &
228,741 & 229,401 & 229,731 & 230,226 \\
\midrule

\multirow{4}{*}{PatchTST~\cite{nie2022time}} & \multirow{4}{*}{Transformer}
& MSE &
0.05093 & \textcolor{blue}{0.06559} & \textcolor{blue}{0.06896} & 0.11279 &
0.07386 & \textcolor{blue}{0.09787} & \textcolor{blue}{0.11675} & 0.146 &
\textcolor{blue}{0.0811} & 0.10967 & \textcolor{blue}{0.14339} & \textcolor{red}{0.17122} \\
& & MAE &
0.20297 & \textcolor{blue}{0.20605} & \textcolor{blue}{0.21076} & 0.25748 &
0.22547 & 0.24022 & 0.26153 & 0.27372 &
\textcolor{blue}{0.22884} & 0.25374 & 0.28907 & 0.29024 \\
& & $R^2$ &
0.98438 & \textcolor{blue}{0.97988} & \textcolor{blue}{0.97885} & 0.96542 &
0.97735 & \textcolor{blue}{0.96999} & \textcolor{blue}{0.9642} & 0.95525 &
\textcolor{blue}{0.97513} & 0.96637 & \textcolor{blue}{0.95605} & \textcolor{blue}{0.94373} \\
& & Params &
101,637 & 102,297 & 102,627 & 103,122 &
101,637 & 102,297 & 102,627 & 103,122 &
101,637 & 102,297 & 102,627 & 103,122 \\
\midrule

\multirow{4}{*}{\textbf{ReLaMix (Ours)}} & \multirow{4}{*}{\textbf{Mixer}}
& MSE &
\textcolor{red}{0.02928} & \textcolor{red}{0.03537} & \textcolor{red}{0.0446} & \textcolor{red}{0.04535} &
\textcolor{red}{0.03218} & \textcolor{red}{0.0477} & \textcolor{red}{0.05176} & \textcolor{red}{0.06314} &
\textcolor{red}{0.05458} & \textcolor{red}{0.08843} & \textcolor{red}{0.11387} & \textcolor{blue}{0.17428} \\
& & MAE &
\textcolor{red}{0.06109} & \textcolor{red}{0.06525} & \textcolor{red}{0.08058} & \textcolor{red}{0.07723} &
\textcolor{red}{0.06421} & \textcolor{red}{0.08656} & \textcolor{red}{0.09153} & \textcolor{red}{0.11175} &
\textcolor{red}{0.09452} & \textcolor{red}{0.13301} & \textcolor{red}{0.16316} & \textcolor{red}{0.28758} \\
& & $R^2$ &
\textcolor{red}{0.99103} & \textcolor{red}{0.98917} & \textcolor{red}{0.98634} & \textcolor{red}{0.98611} &
\textcolor{red}{0.99014} & \textcolor{red}{0.98538} & \textcolor{red}{0.98414} & \textcolor{red}{0.98066} &
\textcolor{red}{0.98327} & \textcolor{red}{0.9729} & \textcolor{red}{0.96511} & \textcolor{red}{0.94752} \\
& & Params &
\textcolor{red}{13,933} & \textcolor{red}{14,593} & \textcolor{red}{14,923} & \textcolor{red}{15,418} &
\textcolor{red}{13,933} & \textcolor{red}{14,593} & \textcolor{red}{14,923} & \textcolor{red}{15,418} &
\textcolor{red}{13,933} & \textcolor{red}{14,593} & \textcolor{red}{14,923} & \textcolor{red}{15,418} \\
\midrule

\multirow{4}{*}{ReLaMix w/o Compression} & \multirow{4}{*}{Ablation}
& MSE &
0.0307 & 0.03357 & 0.03882 & 0.04399 &
0.03917 & 0.04471 & 0.05571 & 0.06373 &
0.04074 & 0.0641 & 0.07684 & 0.11839 \\
& & MAE &
0.05195 & 0.06149 & 0.0666 & 0.08041 &
0.06952 & 0.07942 & 0.09549 & 0.10964 &
0.07253 & 0.11147 & 0.12523 & 0.16515 \\
& & $R^2$ &
0.99058 & 0.9897 & 0.98809 & 0.98651 &
0.988 & 0.9863 & 0.98293 & 0.98048 &
0.98752 & 0.98036 & 0.97646 & 0.96372 \\
& & Params &
45,453 & 46,113 & 46,443 & 46,938 &
45,453 & 46,113 & 46,443 & 46,938 &
45,453 & 46,113 & 46,443 & 46,938 \\
\midrule

\multirow{4}{*}{ReLaMix w/o Residual} & \multirow{4}{*}{Ablation}
& MSE &
0.23709 & 0.22368 & 0.22027 & 0.23014 &
0.21815 & 0.23976 & 0.2538 & 0.26705 &
0.23145 & 0.24142 & 0.27429 & 0.30958 \\
& & MAE &
0.32774 & 0.31816 & 0.30261 & 0.33219 &
0.32887 & 0.31861 & 0.34351 & 0.35568 &
0.33553 & 0.33762 & 0.36544 & 0.39962 \\
& & $R^2$ &
0.92729 & 0.9314 & 0.93245 & 0.92942 &
0.93309 & 0.92647 & 0.92217 & 0.91811 &
0.929 & 0.92595 & 0.91588 & 0.90507 \\
& & Params &
15,853 & 16,513 & 16,843 & 17,338 &
15,853 & 16,513 & 16,843 & 17,338 &
15,853 & 16,513 & 16,843 & 17,338 \\
\bottomrule
\end{tabular}
}
\end{table*}

\subsection{Multi-Level Skip Connections}

Deep mixing networks may suffer from degraded gradient propagation as depth increases, especially when modeling volatile and latency-degraded financial streams. To enhance training stability and promote robust information flow across layers, ReLaMix incorporates \emph{multi-level skip refinement} through cumulative residual connections.

Specifically, at the $l$-th layer, we aggregate transformed representations from all preceding layers to form an additional skip signal:

\vspace{-2.1em}
\begin{equation}
\mathbf{S}^{(l)} = \alpha \sum_{j=0}^{l-1} g_j\!\left(\mathbf{H}^{(j)}\right),
\vspace{-0.6em}
\end{equation}

where $g_j(\cdot)$ denotes a lightweight linear projection, and $\alpha$ is a scaling coefficient (set to $0.1$ in our implementation). The layer update is then defined as:

\vspace{-1.2em}
\begin{equation}
\mathbf{H}^{(l)} = \mathrm{MixBlock}\!\left(\mathbf{H}^{(l-1)}\right) + \mathbf{S}^{(l)}.
\vspace{-0.6em}
\end{equation}

This dense residual pathway acts as an information highway, allowing low-level predictive cues to directly reach deeper layers without being attenuated by intermediate bottleneck processing. Consequently, ReLaMix achieves faster convergence and improved robustness under ZOH-induced observation staleness.

\subsection{Training Objective}

ReLaMix is trained end-to-end in a \emph{corrupted-to-clean} manner, learning to recover future market dynamics from ZOH-corrupted histories. Given an input window $\mathbf{X}^{obs}$, the network predicts $\hat{\mathbf{Y}} \in \mathbb{R}^{k \times D_{out}}$ over the next $k$ steps, supervised by the corresponding clean target sequence $\mathbf{Y}$.

We optimize the model parameters by minimizing the mean squared error (MSE) loss:
\vspace{-1em}
\begin{equation}
\mathcal{L} = \frac{1}{N}\sum_{i=1}^{N}
\left\|\hat{\mathbf{Y}}_i - \mathbf{Y}_i\right\|_{F}^{2},
\vspace{-0.6em}
\end{equation}

where $N$ denotes the number of training samples. This objective encourages ReLaMix to accurately reconstruct the underlying continuous financial dynamics, while suppressing the persistence of latency-induced ZOH artifacts in the predicted trajectories.

\section{EXPERIMENTS}

In this section, we evaluate the proposed ReLaMix framework under a simulated latency setting with ZOH-corrupted observations. We compare ReLaMix against a broad set of state-of-the-art time-series forecasting models, including strong mixer-based baselines such as TimeMixer. Our experiments assess performance from two complementary perspectives: (i) recovery and forecasting accuracy across varying delay ratios, and (ii) computational efficiency in terms of parameter footprint and inference cost. Overall, the results validate the effectiveness of residual bottleneck mixing for robust second-resolution financial forecasting under realistic delayed observation scenarios.

\subsection{Experimental Setup}

We conduct experiments on high-frequency PAXGUSDT trading data collected from the Binance exchange. The dataset consists of multivariate OHLCV features sampled at \emph{1-second} resolution. Following chronological order, the full series is split into training (70\%), validation (15\%), and testing (15\%) subsets. To further examine model robustness, we additionally report generalization results on BTCUSDT under the same evaluation protocol. 

To simulate realistic transmission latency, we inject random stagnation events into the historical inputs: when a time step is flagged as stagnant, its observation is held constant by repeating the previous value under a ZOH mechanism, producing step-function artifacts and information staleness. Given a corrupted history window of length $L=20$, the model aims to recover the corresponding clean future sequence over a prediction horizon of $k=1, 5, 7, 10$ steps.

All models are implemented in PyTorch and trained end-to-end using MSE loss with the Adam optimizer (learning rate $1\times10^{-3}$) and batch size 32. Experiments are conducted on a single NVIDIA GeForce RTX 4090 GPU, and early stopping based on validation loss is applied to prevent overfitting.

\begin{table}[t]
\centering
\caption{SOTA results on \textbf{BTCUSDT} under simulated ZOH delays. 
Best and second-best results are highlighted in \textcolor{red}{red} and 
\textcolor{blue}{blue}.}
\label{tab:sota_btc_filtered}
\setlength{\tabcolsep}{4pt}
\renewcommand{\arraystretch}{1.05}
\resizebox{\columnwidth}{!}{
\begin{tabular}{l|c|c|cc|cc}
\toprule
\multirow{2}{*}{\textbf{Model}} &
\multirow{2}{*}{\textbf{Type}} &
\multirow{2}{*}{\textbf{Metric}} &
\multicolumn{2}{c|}{\textbf{Delay 25\%}} &
\multicolumn{2}{c}{\textbf{Delay 35\%}} \\
\cline{4-7}
& & & $k{=}5$ & $k{=}10$ & $k{=}5$ & $k{=}10$ \\
\midrule

\multirow{3}{*}{LSTM~\cite{hochreiter1997long}} & \multirow{3}{*}{RNN}
& MSE & 0.0332 & 0.0307 & \textcolor{blue}{0.0334} & 0.0305 \\
& & MAE & 0.0287 & \textcolor{blue}{0.0297} & 0.0307 & 0.0302 \\
& & $R^2$ & 0.7700 & 0.7831 & \textcolor{blue}{0.7681} & 0.7845 \\
\midrule

\multirow{3}{*}{GRU~\cite{dey2017gate}} & \multirow{3}{*}{RNN}
& MSE & 0.0345 & 0.0309 & 0.0345 & 0.0306 \\
& & MAE & 0.0346 & 0.0341 & 0.0351 & 0.0305 \\
& & $R^2$ & 0.7610 & 0.7814 & 0.7609 & 0.7832 \\
\midrule

\multirow{3}{*}{TimeMixer~\cite{wang2024timemixer}} & \multirow{3}{*}{Mixer}
& MSE & \textcolor{blue}{0.0329} & 0.0308 & 0.0342 & 0.0310 \\
& & MAE & \textcolor{blue}{0.0240} & 0.0309 & 0.0386 & 0.0323 \\
& & $R^2$ & \textcolor{blue}{0.7722} & 0.7819 & 0.7631 & 0.7806 \\
\midrule

\multirow{3}{*}{TimesNet~\cite{wu2022timesnet}} & \multirow{3}{*}{CNN}& MSE & 0.0334 & \textcolor{blue}{0.0305} & 0.0341 & 0.0308 \\
& & MAE & 0.0277 & 0.0343 & 0.0300 & 0.0323 \\
& & $R^2$ & 0.7684 & \textcolor{blue}{0.7841} & 0.7639 & 0.7819 \\
\midrule

\multirow{3}{*}{PatchTST~\cite{nie2022time}} & \multirow{3}{*}{Transformer}
& MSE & 0.0340 & 0.0308 & 0.0336 & \textcolor{blue}{0.0302} \\
& & MAE & 0.0343 & 0.0323 & \textcolor{blue}{0.0298} & \textcolor{blue}{0.0267} \\
& & $R^2$ & 0.7644 & 0.7819 & 0.7668 & \textcolor{blue}{0.7864} \\
\midrule

\multirow{3}{*}{\textbf{ReLaMix (Ours)}} & \multirow{3}{*}{\textbf{Mixer}}
& MSE & \textcolor{red}{0.0328} & \textcolor{red}{0.0302} & \textcolor{red}{0.0331} & \textcolor{red}{0.0299} \\
& & MAE & \textcolor{red}{0.0226} & \textcolor{red}{0.0293} & \textcolor{red}{0.0234} & \textcolor{red}{0.0222} \\
& & $R^2$ & \textcolor{red}{0.7727} & \textcolor{red}{0.7865} & \textcolor{red}{0.7707} & \textcolor{red}{0.7882} \\

\bottomrule
\end{tabular}
}
\end{table}

\subsection{Comparison with State-of-the-Art Methods}

Table~\ref{tab:sota_ablation_paxg_metric} reports the performance comparison on the second-resolution PAXGUSDT benchmark under simulated ZOH delays. Overall, ReLaMix consistently achieves the best forecasting accuracy across all delay ratios and prediction horizons, while maintaining the smallest parameter footprint among all competing methods. As further visualized in Fig.~\ref{fig:kline_volume}, ReLaMix can effectively recover fine-grained K-line dynamics and volume variations under ZOH corruption, highlighting its capability to suppress stagnation artifacts while preserving informative market fluctuations.

\subsubsection{Accuracy under ZOH latency.}
ReLaMix achieves substantial error reductions over strong baselines. 
At delay ratio $15\%$ and horizon $k=1$, it reaches an MSE of $\mathbf{0.02928}$, outperforming TimesNet (0.04904) and TimeMixer (0.07898). 
This advantage remains consistent under higher delays (25\% and 35\%), indicating robust recovery from stale observations. 
Moreover, ReLaMix attains the lowest MAE and highest $R^2$ across all settings, confirming accurate signal reconstruction.

\subsubsection{Efficiency and practical deployment.}
Beyond accuracy, ReLaMix is highly compact, requiring only $13$--$15$K parameters, which is markedly smaller than TimeMixer ($\sim$41K), PatchTST ($\sim$102K), and TimesNet ($\sim$229K). 
This lightweight design is particularly desirable for latency-sensitive forecasting in high-frequency trading environments, where computational overhead directly constrains real-time applicability.

\subsubsection{Summary.}
These results validate that residual bottleneck mixing provides an effective balance between forecasting fidelity and inference efficiency, enabling state-of-the-art high-frequency financial prediction under realistic delayed observation scenarios.

\subsection{Ablation Study}

To verify the effectiveness of the proposed components in ReLaMix, we conduct an ablation study by removing the two key modules: bottleneck compression and residual refinement. The results are reported in Table~\ref{tab:sota_ablation_paxg_metric}. When the compression module is disabled (\emph{ReLaMix w/o Compression}), the parameter count increases substantially from $\sim$14K to $\sim$46K, and forecasting accuracy consistently degrades across all delay ratios, indicating that explicit bottleneck projection is crucial for suppressing ZOH-induced redundancy and filtering stale observations in a compact latent space. More notably, removing residual refinement (\emph{ReLaMix w/o Residual}) leads to a dramatic performance collapse: for example, at delay ratio $15\%$ and horizon $k=1$, the MSE rises sharply from $0.02928$ to $0.23709$, with similarly severe degradation under larger delays and longer horizons. This confirms that residual pathways are essential for preserving predictive market dynamics and preventing information loss under aggressive ZOH stagnation. Overall, the full ReLaMix model, which integrates both compression and residual mixing, achieves the best trade-off between accuracy and efficiency, attaining the lowest errors and highest $R^2$ while maintaining the smallest parameter footprint. These findings validate our design hypothesis that bottleneck compression removes latency artifacts, whereas residual refinement ensures robust signal recovery under realistic delayed observation scenarios.

\subsection{Generalization Study}

To further evaluate the cross-asset generalization ability of ReLaMix, we report additional results on the BTCUSDT benchmark under more challenging ZOH delay settings (25\% and 35\%) with longer prediction horizons ($k=5$ and $k=10$). As shown in Table~\ref{tab:sota_btc_filtered}, ReLaMix consistently achieves the best performance across all evaluated metrics. In particular, under delay ratio 35\% and horizon $k=10$, ReLaMix attains the lowest MSE ($0.0299$) and MAE ($0.0222$), together with the highest $R^2$ score ($0.7882$), outperforming strong mixer and Transformer baselines such as TimeMixer and PatchTST. These results demonstrate that the proposed residual bottleneck mixing mechanism not only excels on the main PAXGUSDT benchmark, but also transfers effectively to highly volatile cryptocurrency markets, confirming its robustness under realistic delayed observation scenarios.
\section{Conclusion}

In this paper, we studied high-frequency financial forecasting under realistic latency-induced observation staleness, where historical signals are corrupted by a ZOH mechanism. Unlike standard missing-value settings, ZOH stagnation introduces repeated stepwise artifacts that can mislead forecasting models and induce representational redundancy.

To address this challenge, we proposed \textbf{ReLaMix} (Residual Latency-Aware Mixing Network), a lightweight framework integrating learnable bottleneck compression with residual refinement. By suppressing redundancy from stale observations while preserving informative market dynamics, ReLaMix achieves a strong balance between recovery accuracy and computational efficiency.

Experiments on the second-resolution \textbf{PAXGUSDT} benchmark show that ReLaMix consistently achieves state-of-the-art results across delay ratios and prediction horizons, attaining the lowest MSE/MAE and highest $R^2$ with only $\sim$14K parameters. Furthermore, evaluations on \textbf{BTCUSDT} confirm robust cross-asset generalization even under severe delays (e.g., 35\% ZOH corruption).

Overall, ReLaMix provides an efficient solution for forecasting with delayed and partially stale observations. Future work will extend the framework to more complex latency patterns and explore deployment on hardware-efficient platforms for ultra-low-latency trading.



\bibliographystyle{IEEEtran}  
\bibliography{references}     
\vspace{12pt}
\color{red}

\end{document}